\documentclass{article} 
\usepackage{colm}
\usepackage[colorlinks = true,
            linkcolor = blue,
            urlcolor  = blue,
            citecolor = blue,
            anchorcolor = blue]{hyperref}   
\usepackage[utf8]{inputenc} 
\usepackage[T1]{fontenc}    
\usepackage{url}            
\usepackage{booktabs}       
\usepackage{amsfonts}       
\usepackage{nicefrac}       
\usepackage{microtype}      
\usepackage{xcolor}         

\usepackage{algorithm}
\usepackage{algorithmic}

\usepackage{booktabs}
\usepackage{multirow}
\usepackage{tabularx}
\usepackage{makecell}
\usepackage{amssymb}
\usepackage{pifont}
\usepackage{amsmath}
\usepackage{tcolorbox}
\usepackage{listings}
\usepackage{xcolor}
\tcbuselibrary{skins}
\usepackage{hyperref}
\usepackage{float}

\usepackage{fontawesome}

\usepackage{bbding}
\makeatletter
  \newcommand\figcaption{\def\@captype{figure}\caption}
  \newcommand\tabcaption{\def\@captype{table}\caption}
\makeatother

\usepackage[utf8]{inputenc} 
\usepackage[T1]{fontenc}    
\usepackage{url}            
\usepackage{booktabs}       
\usepackage{amsfonts}       
\usepackage{nicefrac}       
\usepackage{microtype}      
\usepackage{array}
\newcolumntype{C}[1]{>{\centering\arraybackslash}m{#1}}
\usepackage{xcolor}         
\usepackage{color, colortbl}
\definecolor{citecolor}{HTML}{2980b9}
\definecolor{linkcolor}{HTML}{c0392b}
\definecolor{darkorange}{HTML}{FF8C00}
\definecolor{chocolate}{HTML}{D2691E}
\definecolor{darkgreen}{HTML}{006400}
\definecolor{darkblue}{HTML}{00008B}
\definecolor{mediumblue}{HTML}{0000CD}
\definecolor{dodgerblue}{HTML}{1E90FF}
\definecolor{royalblue}{HTML}{4169E1}
\definecolor{shadecolor}{RGB}{237,237,237}
\definecolor{backred}{RGB}{255, 190, 190}
\definecolor{backblue}{RGB}{210, 230, 250}

\definecolor{zrrgreen}{HTML}{008000}
\definecolor{zrrblue}{HTML}{4682B4}
\definecolor{zrrred}{HTML}{B22222}

\usepackage{amsmath}
\usepackage{amssymb}
\usepackage{graphicx}

\usepackage{multirow}
\usepackage{makecell}
\usepackage{caption}

\usepackage{adjustbox}
\usepackage{wrapfig}

\usepackage{float}
\usepackage{subfig}

\usepackage{tikz}
\usepackage{tcolorbox}
\tcbuselibrary{skins}

\usepackage{longtable}

\usepackage[nosfdefault]{comicneue} 
\usepackage{booktabs}       
\usepackage{tabularx}
\usepackage{makecell}
\usepackage{array}

\newcolumntype{C}[1]{>{\centering\arraybackslash}m{#1}}

\usepackage{listings}
\usepackage{subcaption}
\usepackage{CJKutf8}
\usepackage{wasysym}
\newcommand{\huggingface}{\raisebox{-1.5pt}{\includegraphics[height=1.05em]{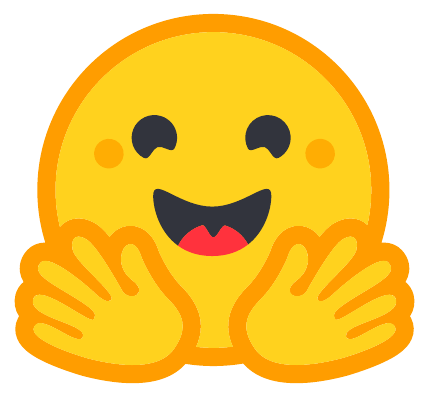}}\xspace}
\newcommand{\github}{\raisebox{-1.5pt}{\includegraphics[height=1.05em]{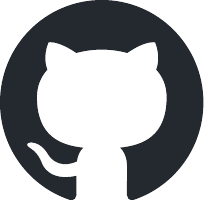}}\xspace}


\usepackage{framed}
\usepackage{makecell}
\usepackage{listings}
\definecolor{lightgray}{rgb}{.9,.9,.9}
\definecolor{darkgray}{rgb}{.4,.4,.4}
\definecolor{purple}{rgb}{0.65, 0.12, 0.82}
\lstdefinelanguage{JavaScript}{
  keywords={break, case, catch, continue, debugger, default, delete, do, else, false, finally, for, function, if, in, instanceof, new, null, return, switch, this, throw, true, try, typeof, var, void, while, with},
  morecomment=[l]{//},
  morecomment=[s]{/*}{*/},
  morestring=[b]',
  morestring=[b]",
  ndkeywords={class, export, boolean, throw, implements, import, this},
  keywordstyle=\color{blue}\bfseries,
  ndkeywordstyle=\color{darkgray}\bfseries,
  identifierstyle=\color{black},
  commentstyle=\color{purple}\ttfamily,
  stringstyle=\color{red}\ttfamily,
  sensitive=true
}
\usepackage{xspace}
\usepackage[shortlabels]{enumitem}

\newcommand\blfootnote[1]{%
  \begingroup
  \renewcommand\thefootnote{}\footnote{#1}%
  \addtocounter{footnote}{-1}%
  \endgroup
}

\usepackage{wrapfig}
\usepackage{float}
\usepackage{xcolor}
\definecolor{kc}{rgb}{0.09, 0.45, 0.27}

\definecolor{lighttan}{rgb}{0.97,0.90,0.85}

\definecolor{CreamyBrown1}{HTML}{DBCCBD}
\definecolor{CreamyBrown2}{HTML}{F6EEE1}

\title{\textit{UniDoc-RL}: Coarse-to-Fine Visual RAG with Hierarchical Actions and Dense Rewards}

\author{Jun Wang$^{*}$, Shuo Tan$^{*}$, Zelong Sun$^{*}$, Tiancheng Gu, Yongle Zhao, Ziyong Feng \\ \textbf{Kaicheng Yang$^{\ddagger}$, Zhiwu Lu$^{\ddagger}$}\\
\\
 \github \textbf{GitHub}  {\url{https://github.com/deepglint/UniDoc-RL}} \\
\huggingface  \textbf{HuggingFace} \url{https://huggingface.co/datasets/DeepGlint-AI/UniDoc-RL}
}

\colmfinalcopy 

\begin{document}

\maketitle

\blfootnote{$^{*}$ Equal Contribution. $^{\ddagger}$ Corresponding Author.} 

\vspace{-0.5cm}

\begin{abstract}
Retrieval-Augmented Generation (RAG) extends Large Vision-Language Models (LVLMs) with external visual knowledge. However, existing visual RAG systems typically rely on generic retrieval signals that overlook the fine-grained visual semantics essential for complex reasoning. To address this limitation, we propose \textbf{UniDoc-RL}, a unified reinforcement learning framework in which an LVLM agent jointly performs retrieval, reranking, active visual perception, and reasoning.
UniDoc-RL formulates visual information acquisition as a sequential decision-making problem with a hierarchical action space. Specifically, it progressively refines visual evidence from coarse-grained document retrieval to fine-grained image selection and active region cropping, allowing the model to suppress irrelevant content and attend to information-dense regions. For effective end-to-end training, we introduce a dense multi-reward scheme that provides task-aware supervision for each action. Based on Group Relative Policy Optimization (GRPO), UniDoc-RL aligns agent behavior with multiple objectives without relying on a separate value network. To support this training paradigm, we curate a comprehensive dataset of high-quality reasoning trajectories with fine-grained action annotations. Experiments on three benchmarks demonstrate that UniDoc-RL consistently surpasses state-of-the-art baselines, yielding up to 17.7\% gains over prior RL-based methods.
\end{abstract}


\section{Introduction}
Retrieval-Augmented Generation (RAG)\citep{guu2020retrieval, yu2023augmentation, yao2022react} improves Large Language Models (LLMs)\citep{bai2025qwen3} by grounding generation in external knowledge. Recent advances in multimodal document understanding extend this paradigm to the visual domain, allowing Large Vision-Language Models (LVLMs)\citep{bai2025qwen25, hurst2024gpt} to reason over external visual evidence\citep{yu2024visrag, faysse2024colpali}. Unlike text, visual documents (\textit{e.g.}, charts and scanned reports) are highly dense and contain substantial redundant background noise, making visual RAG fundamentally more difficult. More recently, several studies~\citep{jin2025search, wang2025vrag} apply Reinforcement Learning (RL) after supervised fine-tuning to further improve visual RAG performance. Nevertheless, developing effective visual RAG systems remains challenging.

\begin{figure*}[t!]
    \centering
    \includegraphics[width=0.95\linewidth]{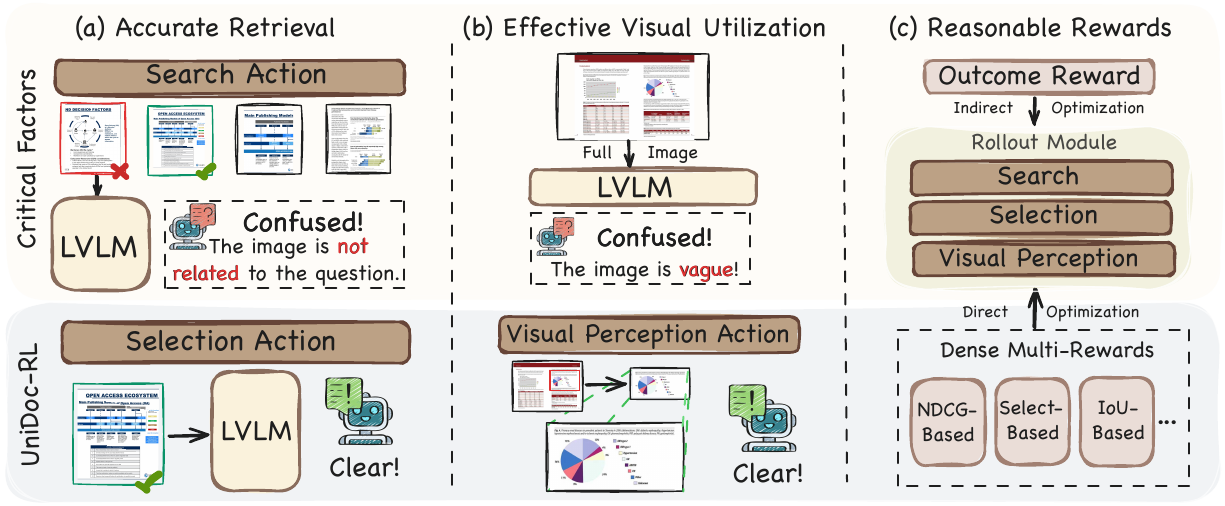}
    \vspace{-2.5mm}
    \caption{\textbf{Three critical factors for Visual RAG.} UniDoc-RL address these challenges through the (a) \textit{Precise Selection} action to bridge the semantic gap between coarse retrieval and reasoning, and an (b) \textit{Active Visual Perception} action to focus on information-dense regions, both optimized via a (c) \textit{Dense multi-reward} mechanism.}
    \label{fig:overview_intro}
    \vspace{-8mm}
\end{figure*}

As illustrated in Figure~\ref{fig:overview_intro}, the success of a visual RAG system hinges on three critical factors: (1) \textbf{Accurate Retrieval}: Erroneous retrieval introduces irrelevant visual context, leading to hallucinations and answer failures; (2) \textbf{Effective Visual Utilization}: Given that images contain dense and often redundant semantic information, the ability to filter noise and focus on key regions is paramount for correct guidance; (3) \textbf{Reasonable Rewards} for Optimization: Effective end-to-end optimization necessitates a dense reward formulation that resolves the credit assignment problem across distinct stages, such as retrieval, selection, active cropping, and reasoning. Existing approaches, however, often address these factors in isolation or fail to model their interdependencies, thereby limiting overall system efficacy.

For retrieval, existing methods~\citep{yu2024visrag, wang2025vrag} typically use decoupled architectures and generic similarity scores from off-the-shelf retrievers. While sufficient for coarse filtering, these scores often fail to capture the task-specific semantics required for complex reasoning. They are also usually static and cannot adapt to query semantics or conversational context, which is particularly problematic in multi-turn settings.
For visual utilization, prior approaches~\citep{yu2024visrag, jin2025search} mainly adopt a passive visual consumption paradigm, directly encoding full images into the model context. This design overlooks the hierarchical nature of visual understanding, retains large amounts of irrelevant background content, and wastes context capacity.
For optimization, existing RL-based methods~\citep{jin2025search, narayan2025deepmmsearch} predominantly use sparse rewards defined only on final outcomes. Such supervision provides no explicit credit assignment for intermediate decisions such as retrieval and cropping, making optimization effectively black-box and less effective for improving the model's internal decision process.

To address these limitations, we propose UniDoc-RL, a unified multimodal reinforcement learning framework in which an LVLM agent jointly performs retrieval, selection, active visual perception, and reasoning.
To bridge the gap between generic retrieval signals and task-specific reasoning requirements, UniDoc-RL introduces a hierarchical action space. It first leverages external tools for efficient coarse-grained retrieval, and then applies an LVLM-driven precise selection action to rerank candidates based on fine-grained semantic alignment with the query. This design effectively filters document-level noise before downstream reasoning.
To further improve visual utilization, we introduce an active visual perception action. Rather than passively encoding full images, the agent learns to perform crop and zoom operations that actively localize informative regions and extract high-value visual evidence. This coarse-to-fine perception strategy resembles human visual attention, reduces redundancy, and preserves high-resolution details.
To address sparse supervision, we design a dense multi-reward system that provides stage-specific training signals throughout the decision process. To support this training paradigm, we curate a comprehensive dataset of high-quality reasoning trajectories with fine-grained action annotations, providing a strong foundation for RL training. Experiments on three benchmarks demonstrate that UniDoc-RL consistently outperforms state-of-the-art baselines, delivering up to 17.7\% gains over prior RL-based methods. 

Our main contributions are summarized as follows:
\textbf{(1)} We propose \textbf{UniDoc-RL}, a unified reinforcement learning framework for visual document RAG that jointly models retrieval, reranking, active visual perception, and reasoning within a single decision process.
\textbf{(2)} We \textbf{build and release a high-quality dataset} of diverse reasoning trajectories with fine-grained action annotations, providing a valuable resource for future research on reinforcement learning for visual RAG.
\textbf{(3)} We \textbf{conduct extensive experiments on three benchmarks} to demonstrate that UniDoc-RL consistently outperforms state-of-the-art baselines.

\section{Related Work}
\noindent{\textbf{Vision-based Retrieval-Augmented Generation.}}
RAG has emerged as a dominant paradigm for knowledge-intensive tasks~\citep{lewis2020retrieval,gao2023retrieval,chen2024benchmarking}. While traditional text-based RAG focuses on interacting with search engines~\citep{wu2025unfolding, chen2024mindsearch,chen2024agent}, the proliferation of digital documents has shifted focus toward multimodal RAG. Early visual RAG approaches, such as ColPali~\citep{faysse2024colpali} and VisRAG~\citep{yu2024visrag}, primarily rely on embedding-based retrieval to align textual queries with visual documents. More recent works have evolved into agentic frameworks~\citep{wang2025vidorag,cho2024m3docrag,jiang2024mmsearch}, utilizing external tools for more precise information extraction. Notably, VRAG-RL~\citep{wang2025vrag} introduces reinforcement learning to incorporate visual perception actions into the RAG pipeline.
In this work, we further introduce a ``Search-Select-Perceive'' coarse-to-fine action space, which bridge the semantic gap between generic retrieval and fine-grained reasoning, enabling the model to progressively filter noise and focus on critical visual evidence.

\noindent{\textbf{Reinforcement Learning for Multimodal Reasoning.}}
RL has proven instrumental in enhancing the reasoning capabilities of LLMs~\citep{guo2025deepseek, jaech2024openai, meng2024simpo, schulman2017proximal}. Recently, this success has been extended to LVLMs~\citep{chen2025r1v,meng2025mm,liu2025visual} and large model-driven agents~\citep{ragen}, particularly for complex tasks requiring multi-step interactions~\citep{jiang2025deepretrievalhackingrealsearch,li2025search}. Despite these advancements, existing RL frameworks typically rely on sparse outcome-based rewards, which struggle to effectively guide intermediate steps like retrieval or cropping due to the credit assignment problem.
In this work, we design specific rewards for retrieval relevance, selection accuracy, and cropping precision, ensuring that every stage of the  pipeline is explicitly supervised and synergistically optimized.

\section{UniDoc-RL}
\subsection{Problem Formulation}
\label{sec:problem}

Given a query $Q$ and a large-scale image corpus $\mathcal{C} = \{c_1, c_2, \dots, c_N\}$, where a subset of images contains the evidence required to answer $Q$, our goal is to retrieve the relevant images, identify the key visual evidence, and generate the correct answer $y$. 

We cast this process as a sequential decision-making problem within a Thought-Action-Observation ($T, A, O$) framework.
At each time step $t$, the model acts as an agent parameterized by a policy $\pi_\theta$. Given the initial query $Q$ and the interaction history $H_{t-1}$, the agent generates a thought $T_t$ and an action $A_t$. After executing $A_t$, it receives an observation $O_t$ from the environment. The interaction proceeds iteratively until the agent determines that sufficient information has been collected to generate the final answer.

\begin{figure*}[t!]
    \centering
    \includegraphics[width=0.98\linewidth]{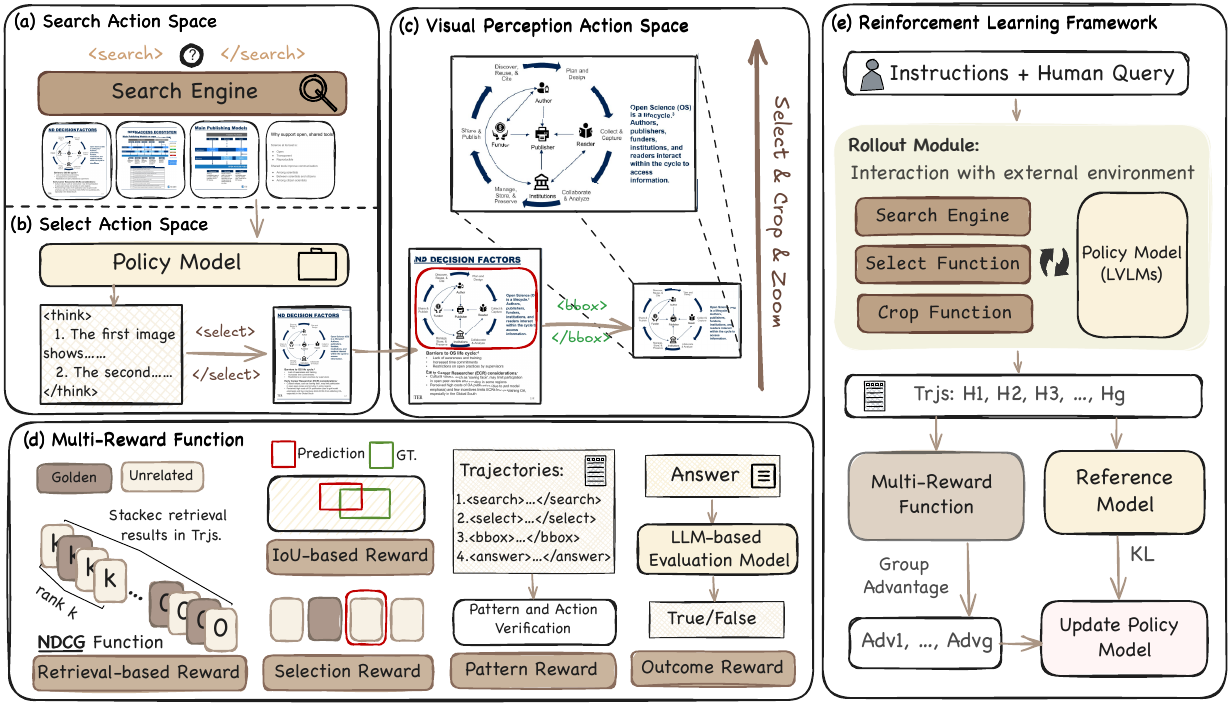}
    \vspace{-2mm}
    \caption{\textbf{Overview of UniDoc-RL.} (a)(b)(c) demonstrates the ``Search-Select-Perceive'' coarse-to-fine action space. (d) is the specially designed reward for UniDoc-RL. (e) shows the interaction process between model and external environment, as well as the implementation of the GRPO algorithm.}
    \vspace{-6mm}
    \label{fig:overview_method}
\end{figure*}

\subsection{Action Definition}
\label{sec:action}

As illustrated in Figure~\ref{fig:overview_method}, UniDoc-RL operates over a hierarchical action space that progressively narrows the visual search space: (1) \textbf{Image Search}, which retrieves a broad set of candidates from the external corpus; (2) \textbf{Precise Selection}, which semantically filters irrelevant images; and (3) \textbf{Visual Perception}, which focuses on fine-grained, information-dense regions within the selected images.

\noindent{\textbf{Image Search Action.}}
The process begins with a coarse-grained retrieval step. At time step $t$, the policy $\pi_{\theta}$ generates an action $A_t$ based on the interaction history $H_{t-1}$. We parse $A_t$ to extract a search query $q$ enclosed by the special tokens \texttt{<search>} and \texttt{</search>}. We then invoke an external retrieval function, denoted by $\text{Search}(\cdot,\cdot)$, to retrieve a set of candidate images from the corpus $\mathcal{C}$ as the observation $O_t$:
\begin{equation}
O_{t} = \text{Search}(q, \mathcal{C})
\end{equation}
where $O_t$ denotes the initial pool of retrieved visual documents.

\noindent{\textbf{Precise Selection Action.}}
External retrieval tools often rely on shallow matching signals and therefore may fail to capture the fine-grained semantics required for complex reasoning. To bridge this gap, UniDoc-RL introduces an LVLM-based selection mechanism. The model evaluates the relevance of each candidate in $O_t$ with respect to the interaction history and generates the next action $A_{t+1}$. Specifically, we extract the indices of relevant images, denoted by $\mathcal{I}$, from the tokens \texttt{<select>} and \texttt{</select>}. The refined observation $O_{t+1}$ is then obtained by filtering the candidate set:
\begin{equation}
O_{t+1} = \text{Select}(O_t, \mathcal{I})
\end{equation}
where $\text{Select}(\cdot,\cdot)$ retains only the images deemed semantically relevant by the LVLM, thereby reducing noise for downstream reasoning.

\noindent{\textbf{Visual Perception Action.}}
Visual scenes often contain dense semantic content, while user queries typically concern only specific local regions. Therefore, we introduce a \textit{Visual Perception} action that enables the model to actively localize and focus on essential information. The model generates an action $A_{t+2}$ that specifies a target region $R$, delimited by the tokens \texttt{<bbox>} and \texttt{</bbox>}. We then apply a perception function, denoted by $\text{VP}(\cdot,\cdot)$, which performs region selection, cropping, and adaptive zooming to extract the region of interest:
\begin{equation}
O_{t+2} = \text{VP}(O_{t+1}, R)
\end{equation}
This operation converts the selected images into high-resolution, query-focused visual observations and effectively removes redundant content.

\subsection{Reward Function Tailored}
\label{sec:reward}
Unlike traditional RL methods that rely solely on sparse outcome-based rewards, UniDoc-RL employs a comprehensive multi-reward mechanism. This dense reward system provides fine-grained feedback at each stage of the pipeline.

\noindent{\textbf{Image Search Reward.}}
High-quality retrieval is the foundation of accurate reasoning. To optimize this process, we employ the Normalized Discounted Cumulative Gain (NDCG) metric. Consider a trajectory containing $m$ retrieval actions, where the $i$-th action returns a set of $k$ candidate images $\mathcal{C}_{cand}^i = \{c_0^i, c_1^i, \ldots, c_k^i\}$. As shown in Figure~\ref{fig:overview_method}, we aggregate these candidates into a unified list $\mathcal{C}_{traj}$ by interleaving them based on their rank:
\begin{equation}
C_{trj} = \{c_0^1, \ldots, c_0^m, c_1^1, \ldots, c_1^m, \ldots, c_k^1, \ldots, c_k^m\}
\end{equation}
We then compute the NDCG score:
\begin{equation}
r_{ir} = NDCG(C_{trj}, C_{rel})
\end{equation}
where $NDCG(\cdot,\cdot)$ denotes the NDCG score function, and $C_{rel}$ is the collection of relevant golden images.


\noindent{\textbf{Precise Selection Reward.}}
After coarse retrieval, the model must select the most relevant image to minimize noise. For the $i$-th retrieval step, let $c_{sel}^i$ be the image selected by the LVLM from $\mathcal{C}_{cand}^i$. The selection reward is defined as:
\begin{equation}
r_{sel}^i = \begin{cases} 
1 & \text{if } c_{sel}^i \in C_{rel} \\
0 & \text{if } c_{sel}^i \notin C_{rel}
\end{cases}
\end{equation}
However, when $\mathcal{C}_{cand}^i$ contains no ground-truth images, the standard reward remains consistently zero regardless of the selection. This results in a non-informative signal that fails to provide discriminative feedback, making it difficult for the model to learn relative relevance among candidates. To address this, we employ a pseudo-supervision strategy: we designate the top-ranked candidate from the initial retrieval, $c_0^i$, as a pseudo-positive target. 
The final selection reward is the average over all $m$ steps: 
\begin{equation}
r_{sel} = \frac{1}{m} \sum_{i=1}^{m} r_{sel}^i.
\end{equation}

\noindent{\textbf{Visual Perception Reward.}}
To encourage accurate localization of key information, we design a reward based on the Intersection over Union (IoU) between predicted crops and ground-truth regions. Let $\hat{\mathcal{B}} = \{\hat{b}_1, \ldots, \hat{b}_M\}$ be the set of predicted bounding boxes and $\mathcal{B}^* = \{b^*_1, \ldots, b^*_N\}$ be the ground-truth boxes. As shown in Figure~\ref{fig:overview_method}, we compute the IoU for each predicted box $\hat{b}_j$ against the best-matching ground-truth box: $r_{crop}=IoU(\mathcal{B}^*,\hat{\mathcal{B}})$



\noindent{\textbf{Pattern Reward.}}
To ensure the structural integrity of the generated thought-action sequences, we introduce a rule-based pattern reward, $r_{pat}$. This reward checks whether the generated trajectory $H$ adheres to the predefined format constraints (\textit{e.g.}, correct usage of XML tags like \texttt{<search>}, \texttt{<select>}, \texttt{<bbox>}). It penalizes malformed outputs, ensuring the model produces valid, executable actions.

\noindent{\textbf{Outcome Reward.}}
The ultimate goal is to generate the correct answer. We employ a model-based outcome reward $r_{ans}$ to evaluate the final response quality:
\begin{equation}
r_{ans} = \mathcal{RM}(A_{gen}, A_{ref}, Q)
\end{equation}
where $\mathcal{RM}$ is a reward model that compares the generated answer $A_{gen}$ with the reference $A_{ref}$. This sparse signal aligns the entire decision process with the final reasoning objective.

The final reward $r_{total}$ is defined as:
\begin{equation}
\label{eq:total_reward}
r_{total} = \lambda_1 r_{pat} + \lambda_2 r_{ir} + \lambda_3 r_{sel} + \lambda_4 r_{crop} + \lambda_5 r_{ans}
\end{equation}
where $\lambda_1, \ldots, \lambda_5$ are hyperparameters to balance the relative importance of each term.

\subsection{Multi-Round Generation with Hierarchical Actions}
\label{sec:generate}
Complex reasoning tasks often require multiple steps of information gathering and refinement. As detailed in Algorithm~\ref{alg:rollout}, UniDoc-RL interacts with the external environment iteratively. In each round, the model generates a thought-action sequence, executes the action, and receives an observation (\textit{e.g.}, retrieved images or cropped regions). Crucially, to align with the LVLM's pre-training data distribution where visual inputs typically appear in user prompts, we insert the observation into the conversation history under the ``User'' role. This design ensures that the model can effectively process the new visual information without disrupting its internal representations.


\begin{algorithm}[!t]
\small 
\caption{Iterative Interaction Process in UniDoc-RL}
\label{alg:rollout}
\begin{algorithmic}[1]
\renewcommand{\algorithmicrequire}{\textbf{Input:}}
\renewcommand{\algorithmicensure}{\textbf{Output:}}
\REQUIRE Input query $Q$, Policy model $\pi_{\theta}$, External environment $\mathcal{E}$, Max steps $T_{max}$.
\ENSURE Final trajectory $H$.
\STATE Initialize history $H \gets \{(Role: User, Content: Q)\}$, step $t \gets 0$
\WHILE{$t < T_{max}$}
    \STATE Generate response $A_t \sim \pi_{\theta}(\cdot \mid H)$
    \STATE Append $(Role: Assistant, Content: A_t)$ to $H$
    
    \IF{\texttt{<search></search>} in $A_t$}
        \STATE Extract search query $q \gets \text{Parse}(A_t)$ and Retrieve candidates $\mathcal{C}_{cand} \gets \text{Search}(q, \mathcal{C})$
        \STATE Observation $O_t \gets \text{Format}(\mathcal{C}_{cand})$
    \ELSIF{\texttt{<select></select>} in $A_t$}
        \STATE Extract index $idx \gets \text{Parse}(A_t)$
        and Select image $I_{sel} \gets \mathcal{C}_{cand}[idx]$
        \STATE Observation $O_t \gets I_{sel}$
    \ELSIF{\texttt{<bbox></bbox>} in $A_t$}
        \STATE Extract box $R \gets \text{Parse}(A_t)$
        and Crop image $I_{vp} \gets \text{VP}(I_{sel}, R)$
        \STATE Observation $O_t \gets I_{vp}$
    \ELSIF{\texttt{<answer></answer>} in $A_t$}
        \STATE \textbf{return} $H$ \COMMENT{Terminate upon answer generation}
    \ENDIF
    
    \STATE Append $(Role: User, Content: O_t)$ to $H$ \COMMENT{Inject observation as user input}
    and $t \gets t + 1$
\ENDWHILE
\STATE \textbf{return} $H$
\end{algorithmic}
\end{algorithm}
\section{Experiments}
\subsection{Training Data Synthesis}
\noindent{\textbf{Data Source.}}
To construct a diverse and comprehensive training corpus, we aggregate samples from several public benchmarks, including SlideVQA~\citep{tanaka2023slidevqa}, Double Bench~\citep{shen2025rightwayassessingdocument}, VisR-Bench~\citep{chen2025visrbenchempiricalstudyvisual}, DocBench~\citep{zou2024docbenchbenchmarkevaluatingllmbased}, and DUDE~\citep{vanlandeghem2023documentunderstandingdatasetevaluation}. We implement a rigorous multi-stage filtering pipeline to ensure high data quality. Consequently, we curate a final dataset comprising 12,621 samples for SFT and 5,537 samples for RL. 

\noindent{\textbf{Automated Trajectory Synthesis.}}
\label{sec:data_gen}    
To acquire high-quality supervision that demonstrates effective coordination between retrieval, selection, and perception, we employ Qwen3-VL-235B~\citep{bai2025qwen3} as a teacher agent to synthesize reasoning trajectories. While the generation of search and selection actions follows the standard interaction loop, the construction of \textit{Visual Perception} actions requires specific handling to ensure precision. Therefore, we introduce an intermediate layout analysis step and use the expert document parsing tool Mineru~\citep{niu2025mineru2} to detect layout elements and generate a set of candidate bounding boxes, which represent potential regions of interest (ROIs). Please refer to the \textbf{Appendix} for more details.

\subsection{Training Pipeline and Implementation Details}
To initialize the model with fundamental reasoning capabilities and ensure strict adherence to the defined action formats, we first perform SFT as a cold start. We conduct SFT on llama-factory~\citep{zheng2024llamafactory}, with full parameter fine-tuning and cosine learning scheduler with a warmup ratio of 0.1. We then employ Group Relative Policy Optimization (GRPO)~\citep{shao2024deepseekmath} to fine-tune the model using the multi-reward system defined in Section~\ref{sec:reward}. We conduct GRPO on verl~\citep{sheng2024hybridflow} with the group size setted to 5. All the experiments are conducted on 8 NVIDIA A100 80G GPUs. Please refer to \textbf{Appendix} for more detailed hyperparameters information.

\subsection{Evaluation Datasets and Baselines}
We evaluate performance on three challenging, visually rich benchmarks: ViDoSeek~\citep{wang2025vidorag}, SlideVQA~\citep{tanaka2023slidevqa}, and MMLongBench~\citep{ma2024mmlongbench}. We employ an evaluation model to assess the correctness of generated answers, which returns a binary score. The overall accuracy rate is computed as the evaluation metric.

For baseline comparisons, following~\citep{wang2025vrag}, we select several representative methods spanning text-based and vision-based approaches: (1) \textbf{Vanilla RAG}~\citep{faysse2024colpali} directly uses the original question as a search query, after which LVLMs perform inference on the retrieved results. (2) \textbf{ReAct}~\citep{yao2022react} follows a think-then-act paradigm, where the model iteratively performs query rewriting, information retrieval, and reasoning. (3) \textbf{Search-R1(-VL)}~\citep{jin2025search} is adapted from the Search-R1 baseline, with experimental settings aligned across all methods to ensure fair comparison. (4) \textbf{VRAG-RL}~\citep{wang2025vrag} serves as our most direct baseline, as it also incorporates a visual perception mechanism within an RL framework.



\begin{table}[!t]
    \centering
    
    \resizebox{1.00\textwidth}{!}{
    \setlength{\aboverulesep}{0pt}
    \setlength{\belowrulesep}{0pt}
    \renewcommand{\arraystretch}{1.2} 

    \begin{tabular}{l|cc|cc|ccccc|c}
    \toprule
    \rowcolor{CreamyBrown1} 
    & \multicolumn{2}{c|}{\textsc{\textbf{SlideVQA}}} 
    & \multicolumn{2}{c|}{\textsc{\textbf{ViDoSeek}}} 
    & \multicolumn{5}{c|}{\textsc{\textbf{MMLongBench}}} 
    & \\ 
    
    \rowcolor{CreamyBrown1} 
    \multirow{-2}{*}{\textsc{\textbf{Method}}} 
    & \textbf{Single-hop} & \textbf{Multi-hop} & \textbf{Extraction} & \textbf{Logic} &
    \textbf{Text} & \textbf{Table} & \textbf{Chart} & \textbf{Figure} & \textbf{Layout} 
    & \multirow{-2}{*}{\textsc{\textbf{Overall}}} \\ 
    
    \midrule
    \multicolumn{11}{c}{$\textit{Qwen2.5-VL-3B-Instruct}$}\\
    \midrule
    {\small \faEyeSlash} Vanilla RAG & 15.1&12.1&8.8&14.3&3.9&5.1&1.7&3.1&2.5 &11.2\\
    {\small \faEyeSlash} ReAct &11.8&9.9&5.3&7.4&6.5&3.7&3.9&5.2&2.5&8.4\\
    {\small \faEyeSlash} Search-R1 &17.5&13.8&13.3&20.7&3.4&3.2&4.5&4.1&6.8&14.1\\
    \midrule
    {\small \faEye} Vanilla RAG   & 19.4 & 12.2 & 10.1 & 17.3 & 2.2 & 4.1 & 5.2 & 4.7 & 4.3 & 13.2 \\
    {\small \faEye} ReAct & 15.7 & 10.9 & 6.7 & 14.2 & 2.7 & 3.6 & 3.4 & 3.1 & 5.1 & 10.9 \\
    {\small \faEye} Search-R1-VL & 26.3 & 20.1 & 20.1 & 29.8 & 8.5 & 7.8 & 7.9 & 9.3 & 7.6 & 21.3 \\
    {\small \faEye} VRAG-RL & 65.3 & 38.6 & 63.1  & 73.8 & 22.7 & 16.1 & 21.9 & 21.4 & 19.5 & 53.5 \\
    \midrule
     \rowcolor{CreamyBrown2} {\small \faEye} \textbf{UniDoc-RL} & \textbf{82.2} & \textbf{63.1} & \textbf{78.6} & \textbf{75.9} & \textbf{43.9} & \textbf{33} & \textbf{27.3} & \textbf{50.7} & \textbf{40.6} & \textbf{71.0} \\
    \midrule
    \multicolumn{11}{c}{$\textit{Qwen2.5-VL-7B-Instruct}$}\\
    \midrule
    {\small \faEyeSlash} Vanilla RAG & 26.1 & 10.6 & 24.7 & 30.9 & 8.5 & 5.4& 11.7 & 4.4 & 3.3 & 20.9 \\
    {\small \faEyeSlash} ReAct & 21.2&13.3&14.3&21.3&5.9&5.1&7.3&5.5&1.7&15.8\\
    {\small \faEyeSlash} Search-R1 &28.4&19.7&20.8&30.6& 9.9&6.0&7.9&10.1&5.9&22.2\\
    \midrule
    {\small \faEye} Vanilla RAG & 29.1 & 17.4 & 26.4 & 41.3 & 13.1 & 14.7 & 15.9 & 4.3 & 7.6 & 24.2 \\
    {\small \faEye} ReAct & 34.8 & 20.4 & 27.5 & 42.1 & 10.1 & 12.4 & 10.2 & 6.2 & 7.1 & 26.9 \\
    {\small \faEye} Search-R1-VL & 48.3 & 42.3 & 40.5 & 50.3 & 19.9 & 13.4 & 12.9 & 11.4 & 10.2 & 37.4 \\
    {\small \faEye} VRAG-RL & 69.3 & 43.1 & 60.6 & 74.8 & 26.1 & 26.3 & 24.8 & 25.9 & 21.2 & 57.1 \\
    \midrule
    \rowcolor{CreamyBrown2} {\small \faEye} \textbf{UniDoc-RL} & \textbf{86.3} & \textbf{68.9} & \textbf{80.3} & \textbf{78.7} & \textbf{51.5} & \textbf{39.6} & \textbf{32.2} & \textbf{50.0} & \textbf{40.6} & \textbf{74.8} \\
    \bottomrule
    \end{tabular}
}
\vspace{-2mm}
 \caption{\textbf{Main Results.} The best performance are marked in bold. SlideVQA and ViDoSeek mainly focus on reasoning type, while MMLongBench focuses on the visual type of reference content. OCR-based ({\small \faEyeSlash}) RAG and purely visual ({\small \faEye}) RAG are evaluated with the same prompt and setting.}
 \label{tab:overall_performance}
\vspace{-6mm}
\end{table}

\subsection{Main Results}
Table~\ref{tab:overall_performance} presents a comparative analysis of UniDoc-RL against existing RAG approaches. We summarize the key observations as follows:
\textbf{(1) Limitations of OCR-based RAG.} 
OCR-based methods, which rely solely on extracted textual content, struggle with visually intensive tasks. This performance gap stems from the inevitable loss of critical visual cues (such as spatial layout and geometric relationships) during the text extraction process.
\textbf{(2) Efficacy of Visual RAG.} 
Compared to OCR-based baselines, purely visual RAG methods demonstrate significant improvements by preserving complete visual information. This validates that LVLMs have evolved to effectively capture both textual semantics and spatial relationships within images, often surpassing the capabilities of traditional OCR pipelines.
\textbf{(3) Superiority of UniDoc-RL.} 
RL-based methods consistently outperform supervised baselines across all datasets. Notably, UniDoc-RL achieves state-of-the-art performance, surpassing the strong competitor VRAG-RL by margins of 17.5\% and 17.7\% on 3B and 7B models, respectively. We attribute this substantial gain to two factors: the \textit{Precise Selection} action, which filters noise to provide a high-quality context, and the \textit{Visual Perception} action, which enables fine-grained attention. Furthermore, the multi-reward mechanism ensures synergistic optimization across all stages.
\textbf{(4) Robustness Across Diverse Tasks.} 
The datasets present distinct challenges: SlideVQA and ViDoSeek require complex multi-hop reasoning, while MMLongBench demands precise attention to fine-grained visual details. UniDoc-RL yields consistently superior performance across diverse benchmarks, demonstrating its robustness and indicating that the proposed multi-reward RL framework effectively enhances both logical reasoning and visual perception.

\section{Ablation Study}
\label{sec:ablation}

\subsection{Impact of Hierarchical Actions.}

\begin{wraptable}{tr}{.5\linewidth}
\vspace{-1mm}
\centering


\setlength{\aboverulesep}{0pt}
\setlength{\belowrulesep}{0pt}
\renewcommand{\arraystretch}{1.2} 

\scalebox{0.78}{
\begin{tabular}{cc|cccc}
    \toprule
    \rowcolor{CreamyBrown1} 
    \multicolumn{2}{c|}{\textsc{\textbf{Action Space}}} &
    \multicolumn{4}{c}{\textsc{\textbf{Accuracy}}} \\ 
    
    \rowcolor{CreamyBrown1} 
    \textbf{Select} &\textbf{VP} & \textbf{SVQA} &\textbf{VDS} &\textbf{MMLB} &\textbf{Overall} \\ 

    
    \midrule
     \ding{55}& \ding{55}& 71.8 & 72.8 & 37.0 & 66.6\\
     \ding{51}& \ding{55}& 76.0 & 76.4 & 37.3 & 70.0\\
    \rowcolor{CreamyBrown2} \ding{51} & \ding{51} & \textbf{77.2} & \textbf{77.4} & \textbf{38.0} & \textbf{71.0}\\
    \bottomrule
    \end{tabular}}
\vspace{-2mm}
\caption{\textbf{Ablation for Different Actions.}}
\vspace{-4mm}
\label{tab:ablation_action}
\end{wraptable}
In Table~\ref{tab:ablation_action}, we isolate the contributions of each component in UniDoc-RL with Qwen2.5-VL-3B-Instruct. We have the following insights:
\textbf{(1) Efficacy of Precise Selection.} 
Incorporating the \textit{Precise Selection} action leads to substantial performance gains. This confirms that the selection action significantly improves the recall of relevant images, effectively bridging the semantic gap between the retriever and the reasoner (as further illustrated in Figure~\ref{fig:hit_rate}).
\textbf{(2) Importance of Visual Perception.} 
The addition of the \textit{Visual Perception} action further boosts performance, which validates necessity of the ``active cropping'' mechanism and the effectiveness of our IoU-based reward in guiding the model to localize key evidence.
\textbf{(3) Task-Specific Adaptability.} 
We observe that the impact of specific modules varies by task nature. The \textit{Selection} action yields the most pronounced improvements on reasoning-heavy datasets (SlideVQA and ViDoSeek), suggesting that context quality is paramount for complex reasoning. \textit{Visual Perception} action provides larger gains on MMLongBench, indicating its critical role in tasks requiring fine-grained visual discrimination.

\subsection{Ablation on Multi-Rewards.}
In Table~\ref{tab:ablation_reward}, we conducted an incremental ablation study for our reward design. We observe the following:
\textbf{(1) Benefit of Dense Supervision.} 
Compared to the baseline variant which relies solely on sparse outcome-based and pattern rewards, our proposed multi-reward framework (Full) achieves consistent performance improvements across all three datasets. This demonstrates that dense, step-wise supervision effectively optimizes the accuracy of intermediate actions, preventing error propagation and guiding the model toward more robust reasoning paths. 
\textbf{(2) Task-Reward Alignment.} 
\begin{wraptable}{tr}{.5\linewidth}
\small
\vspace{-4mm}
\centering

\tabcolsep5pt

\setlength{\aboverulesep}{0pt}
\setlength{\belowrulesep}{0pt}
\renewcommand{\arraystretch}{1.2}

\scalebox{0.95}{
\begin{tabular}{c|cccc}

    \toprule
    \rowcolor{CreamyBrown1} 
     &
    \multicolumn{4}{c}{\textsc{\textbf{Accuracy}}} \\ 
    
    \rowcolor{CreamyBrown1} 
    \multirow{-2}{*}{\textsc{\textbf{REWARD}}} & \textbf{SVQA} &\textbf{VDS} &\textbf{MMLB} &\textbf{Overall} \\

    \midrule
     Vanilla & 75.9 & 76.7 & 36.8 & 69.9\\
     + $r_{ir}$ & 76.4 & 76.2 & 36.8 & 70.1\\
     + $r_{ir}$ + $r_{sel}$ & 76.6 & 77.8 & 36.6 & 70.6\\
    \rowcolor{CreamyBrown2} + $r_{ir}$ + $r_{sel}$ + $r_{vp}$ & \textbf{77.2} & \textbf{77.4} & \textbf{38.0} & \textbf{71.0}\\
    \bottomrule
    \end{tabular}}
\vspace{-2mm}
\caption{\textbf{Ablation on Multi-Rewards.}}
\label{tab:ablation_reward}
\end{wraptable}
Corroborating the findings in Table~\ref{tab:ablation_action}, we 
observe that the inclusion of the \textit{Selection Reward} significantly boosts performance on reasoning-heavy datasets, whereas the \textit{Visual Perception Reward} yields the most notable improvements on MMLongBench. This further confirms that tailoring rewards to specific sub-tasks is essential for diverse reasoning requirements.


\section{Analysis}
\subsection{Selection Action Facilitates Better Retrieval Results.}

\begin{wrapfigure}{r}{0.37\linewidth}
\small
\centering
\vspace{-8mm}
\includegraphics[width=0.95\linewidth]{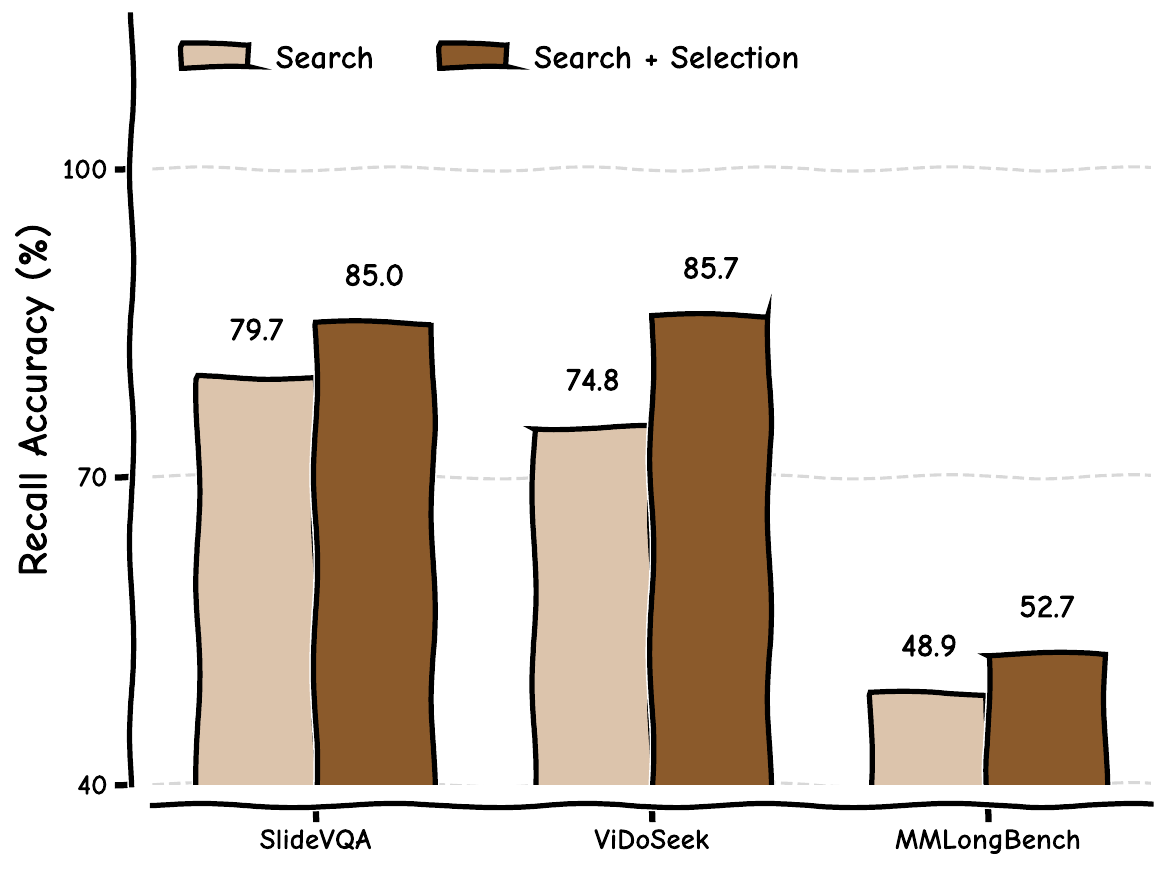}
\vspace{-2mm}
\caption{\textbf{Retrieval Recall} before and after adding the selection action.}
\label{fig:hit_rate}
\vspace{-2mm}
\end{wrapfigure}
UniDoc-RL substantially improves the retrieval hit rate of ground-truth images through the \textit{Precise Selection} action, which helps build a more accurate and informative context for generation. As shown in Figure~\ref{fig:hit_rate}, adding the selection step consistently improves recall over coarse search alone across all three benchmarks. Specifically, recall on SlideVQA and ViDoSeek increases from 79.7\% and 74.8\% to 85.0\% and 85.7\%, respectively, while MMLongBench improves from 48.9\% to 52.7\%. These results show that LVLM-based selection effectively narrows the semantic gap in coarse retrieval by filtering out irrelevant candidates and provides clearer visual evidence for downstream reasoning.

\subsection{SFT Enables Tool Use, While RL Refines the Tool-Selection Behavior}
\label{sec:analysis_behavior}


\begin{wrapfigure}{r}{0.4\linewidth}
\small
\centering
\vspace{-0.1in}
\includegraphics[width=1\linewidth]{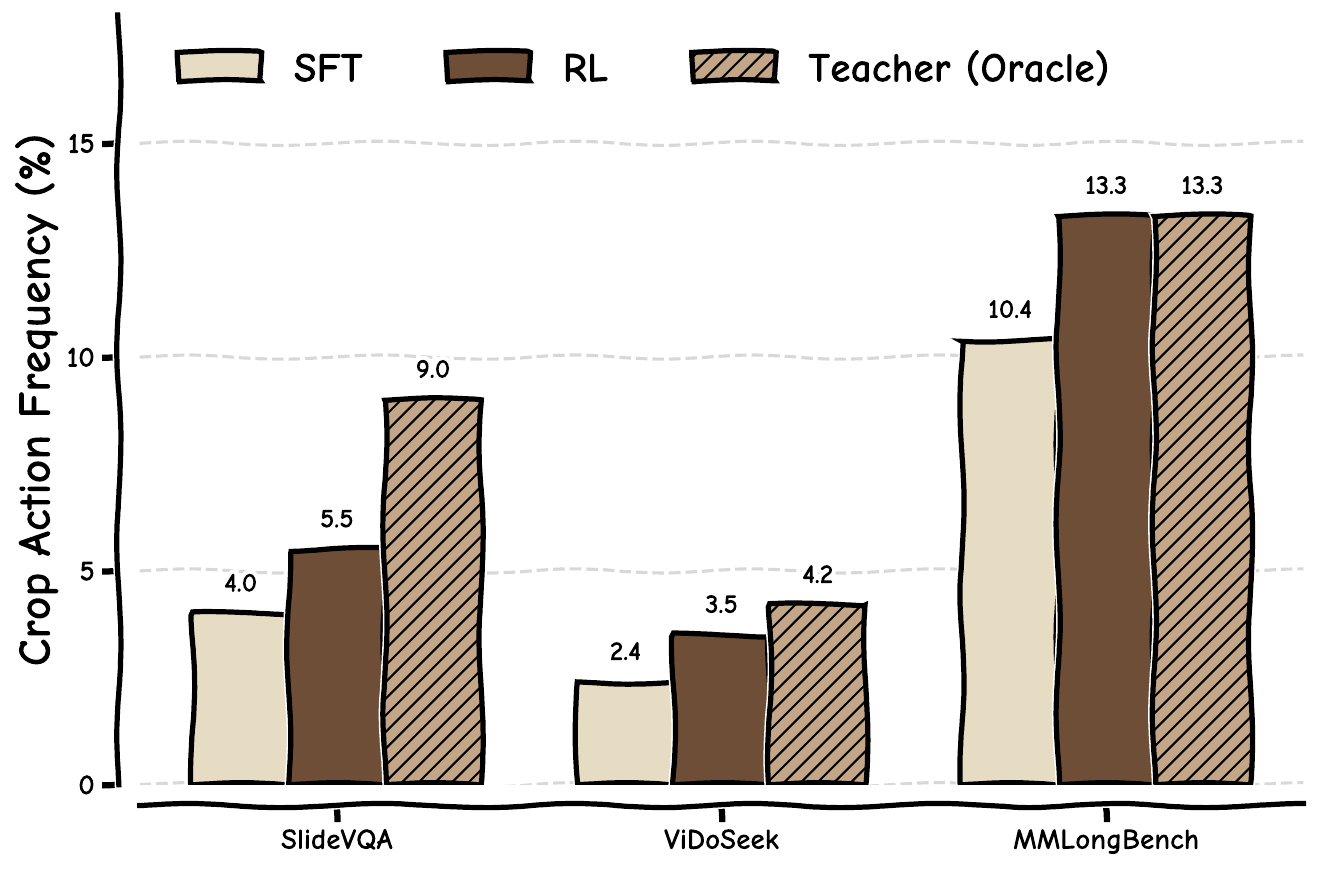}
\vspace{-0.3in}
\caption{\textbf{Visual Perception Action Frequency} across SFT, RL, and Teacher models on three benchmarks.}
\label{fig:crop_frequency}
\vspace{-0.2in}
\end{wrapfigure}

\textbf{RL Encourages Active Information Seeking.} 
As shown in Figure~\ref{fig:crop_frequency}, the SFT model exhibits a conservative strategy with a low crop frequency (\textit{e.g.}, only 2.4\% on ViDoSeek), significantly lagging behind the Teacher model. This suggests that SFT suffers from exposure bias, often defaulting to passive full-image reading. In contrast, UniDoc-RL demonstrates a more active perception pattern. After RL training, the crop frequency increases across all datasets (e.g., rising to 13.3\% on MMLongBench), closely aligning with the Teacher's behavior. Notably, on MMLongBench, which demands fine-grained visual attention, UniDoc-RL matches the Teacher's frequency, indicating it has successfully learned to invoke perception tools when facing detail-intensive queries.


\begin{wrapfigure}{l}{0.5\linewidth}
\small
\centering
\vspace{-4mm}
\includegraphics[width=0.93\linewidth]{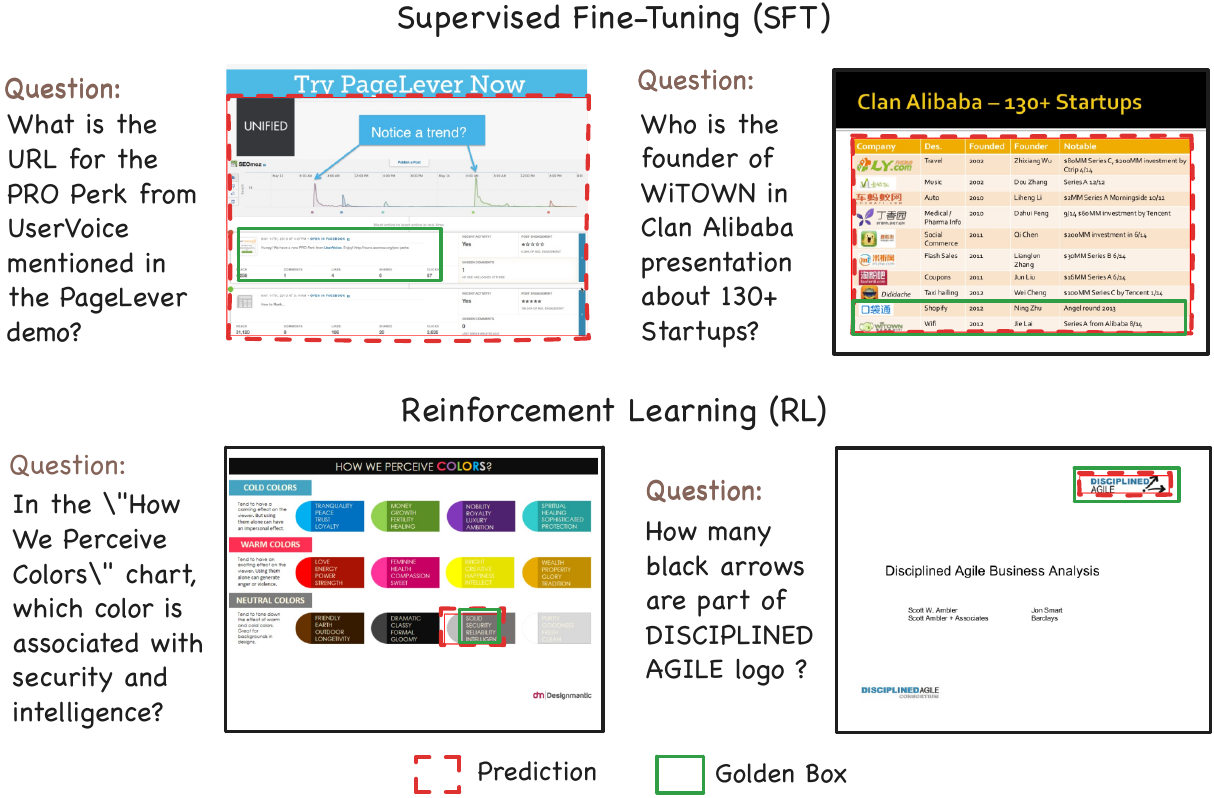}
\vspace{-3mm}
\caption{\textbf{Qualitative Comparison of Visual Perception Action} generated by UniDoc-RL before and after RL fine-tuning.}
\label{fig:crop_analysis}
\vspace{-3mm}
\end{wrapfigure}
\textbf{RL Improves Action Quality and Precision.} 
Beyond the crop frequency, we observe a qualitative shift in how the model crops. As visualized in Figure~\ref{fig:crop_analysis}, the SFT model frequently 
generates``trivial crops'' that encompass the entire image or irrelevant backgrounds, failing to provide additional information gain. Conversely, UniDoc-RL, guided by the IoU-based perception reward and task-specific feedback, learns to execute ``precise crops'' that tightly bound the regions of interest. This shift from coarse, meaningless actions to focused, high-utility perception confirms that our multi-reward RL framework effectively optimizes both the decision to act and the execution of the action.

\subsection{Case Study}
\label{sec:case_study}

Figure~\ref{fig:case_study} in the Appendix presents two representative examples that illustrate UniDoc-RL's adaptive reasoning process. In Case~1, the model retrieves candidate images, selects the one containing relevant training loss curves, and then actively crops a specific sub-region to zoom into a small chart which correctly identifies that the loss spikes around the 150k step. This demonstrates the necessity of the \textit{Visual Perception} action for extracting fine-grained details from complex figures. In Case~2, the model retrieves and selects a table that is already clearly legible, and intelligently skips the crop action, directly extracting the answer. These two cases jointly show that UniDoc-RL learns an adaptive strategy through RL training.

\section{Conclusion}
We present \textbf{UniDoc-RL}, a unified reinforcement learning framework for visual RAG that orchestrates retrieval, reranking, active visual perception, and reasoning within an LVLM agent. Through a hierarchical action space, the model progressively filters visual evidence from coarse retrieval to fine-grained selection and region cropping, effectively suppressing noise and focusing on key information. To overcome the limitations of sparse outcome-based rewards, a dense multi-reward mechanism provides stage-aware supervision, enabling end-to-end optimization via GRPO. Additionally, we curate and publicly release a high-quality dataset of reasoning trajectories with fine-grained action annotations to support future research. Extensive experiments on three benchmarks demonstrate that UniDoc-RL consistently outperforms state-of-the-art baselines.

\clearpage
{
\bibliography{ViDoRAG_R1}

@String(NeurIPS = {Adv. Neural Inform. Process. Syst.})

@String(ICML  = {Int. Conf. Mach. Learn.})

@String(ICLR  = {Int. Conf. Learn. Represent.})

@String(AAAI  = {AAAI})

@String(NeurIPS = {NeurIPS})

@String(ICML  = {ICML})

@String(ICLR  = {ICLR})

@article{chen2024mindsearch,
  title={Mindsearch: Mimicking human minds elicits deep ai searcher},
  author={Chen, Zehui and Liu, Kuikun and Wang, Qiuchen and Liu, Jiangning and Zhang, Wenwei and Chen, Kai and Zhao, Feng},
  journal={arXiv preprint arXiv:2407.20183},
  year={2024}
}

@article{chen2024agent,
  title={Agent-flan: Designing data and methods of effective agent tuning for large language models},
  author={Chen, Zehui and Liu, Kuikun and Wang, Qiuchen and Zhang, Wenwei and Liu, Jiangning and Lin, Dahua and Chen, Kai and Zhao, Feng},
  journal={arXiv preprint arXiv:2403.12881},
  year={2024}
}

@article{wang2025vidorag,
  title={ViDoRAG: Visual Document Retrieval-Augmented Generation via Dynamic Iterative Reasoning Agents},
  author={Wang, Qiuchen and Ding, Ruixue and Chen, Zehui and Wu, Weiqi and Wang, Shihang and Xie, Pengjun and Zhao, Feng},
  journal={arXiv preprint arXiv:2502.18017},
  year={2025}
}

@article{jin2025search,
  title={Search-r1: Training llms to reason and leverage search engines with reinforcement learning},
  author={Jin, Bowen and Zeng, Hansi and Yue, Zhenrui and Yoon, Jinsung and Arik, Sercan and Wang, Dong and Zamani, Hamed and Han, Jiawei},
  journal={arXiv preprint arXiv:2503.09516},
  year={2025}
}

@article{yu2024visrag,
  title={Visrag: Vision-based retrieval-augmented generation on multi-modality documents},
  author={Yu, Shi and Tang, Chaoyue and Xu, Bokai and Cui, Junbo and Ran, Junhao and Yan, Yukun and Liu, Zhenghao and Wang, Shuo and Han, Xu and Liu, Zhiyuan and others},
  journal={arXiv preprint arXiv:2410.10594},
  year={2024}
}

@inproceedings{faysse2024colpali,
  title={Colpali: Efficient document retrieval with vision language models},
  author={Faysse, Manuel and Sibille, Hugues and Wu, Tony and Omrani, Bilel and Viaud, Gautier and Hudelot, C{\'e}line and Colombo, Pierre},
  booktitle={ICLR},
  year={2024}
}

@inproceedings{zheng2024llamafactory,
  title={LlamaFactory: Unified Efficient Fine-Tuning of 100+ Language Models},
  author={Yaowei Zheng and Richong Zhang and Junhao Zhang and Yanhan Ye and Zheyan Luo and Zhangchi Feng and Yongqiang Ma},
  booktitle={Proceedings of the 62nd Annual Meeting of the Association for Computational Linguistics (Volume 3: System Demonstrations)},
  address={Bangkok, Thailand},
  publisher={ACL},
  year={2024},
  url={http://arxiv.org/abs/2403.13372}
}

@article{sheng2024hybridflow,
  title   = {HybridFlow: A Flexible and Efficient RLHF Framework},
  author  = {Guangming Sheng and Chi Zhang and Zilingfeng Ye and Xibin Wu and Wang Zhang and Ru Zhang and Yanghua Peng and Haibin Lin and Chuan Wu},
  year    = {2024},
  journal = {arXiv preprint arXiv: 2409.19256}
}

@inproceedings{tanaka2023slidevqa,
  title={Slidevqa: A dataset for document visual question answering on multiple images},
  author={Tanaka, Ryota and Nishida, Kyosuke and Nishida, Kosuke and Hasegawa, Taku and Saito, Itsumi and Saito, Kuniko},
  booktitle={AAAI},
  pages={13636--13645},
  year={2023}
}

@article{ma2024mmlongbench,
  title={Mmlongbench-doc: Benchmarking long-context document understanding with visualizations},
  author={Ma, Yubo and Zang, Yuhang and Chen, Liangyu and Chen, Meiqi and Jiao, Yizhu and Li, Xinze and Lu, Xinyuan and Liu, Ziyu and Ma, Yan and Dong, Xiaoyi and others},
  journal={arXiv preprint arXiv:2407.01523},
  year={2024}
}

@article{jaech2024openai,
  title={Openai o1 system card},
  author={Jaech, Aaron and Kalai, Adam and Lerer, Adam and Richardson, Adam and El-Kishky, Ahmed and Low, Aiden and Helyar, Alec and Madry, Aleksander and Beutel, Alex and Carney, Alex and others},
  journal={arXiv preprint arXiv:2412.16720},
  year={2024}
}

@article{guo2025deepseek,
  title={Deepseek-r1: Incentivizing reasoning capability in llms via reinforcement learning},
  author={Guo, Daya and Yang, Dejian and Zhang, Haowei and Song, Junxiao and Zhang, Ruoyu and Xu, Runxin and Zhu, Qihao and Ma, Shirong and Wang, Peiyi and Bi, Xiao and others},
  journal={arXiv preprint arXiv:2501.12948},
  year={2025}
}

@article{schulman2017proximal,
  title={Proximal policy optimization algorithms},
  author={Schulman, John and Wolski, Filip and Dhariwal, Prafulla and Radford, Alec and Klimov, Oleg},
  journal={arXiv preprint arXiv:1707.06347},
  year={2017}
}

@article{shao2024deepseekmath,
  title={Deepseekmath: Pushing the limits of mathematical reasoning in open language models},
  author={Shao, Zhihong and Wang, Peiyi and Zhu, Qihao and Xu, Runxin and Song, Junxiao and Bi, Xiao and Zhang, Haowei and Zhang, Mingchuan and Li, YK and Wu, Y and others},
  journal={arXiv preprint arXiv:2402.03300},
  year={2024}
}

@article{Qwen2.5-VL,
  title={Qwen2.5-VL Technical Report},
  author={Bai, Shuai and Chen, Keqin and Liu, Xuejing and Wang, Jialin and Ge, Wenbin and Song, Sibo and Dang, Kai and Wang, Peng and Wang, Shijie and Tang, Jun and Zhong, Humen and Zhu, Yuanzhi and Yang, Mingkun and Li, Zhaohai and Wan, Jianqiang and Wang, Pengfei and Ding, Wei and Fu, Zheren and Xu, Yiheng and Ye, Jiabo and Zhang, Xi and Xie, Tianbao and Cheng, Zesen and Zhang, Hang and Yang, Zhibo and Xu, Haiyang and Lin, Junyang},
  journal={arXiv preprint arXiv:2502.13923},
  year={2025}
}

@article{wu2025unfolding,
  title={Unfolding the Headline: Iterative Self-Questioning for News Retrieval and Timeline Summarization},
  author={Wu, Weiqi and Huang, Shen and Jiang, Yong and Xie, Pengjun and Huang, Fei and Zhao, Hai},
  journal={arXiv preprint arXiv:2501.00888},
  year={2025}
}

@article{cho2024m3docrag,
  title={M3docrag: Multi-modal retrieval is what you need for multi-page multi-document understanding},
  author={Cho, Jaemin and Mahata, Debanjan and Irsoy, Ozan and He, Yujie and Bansal, Mohit},
  journal={arXiv preprint arXiv:2411.04952},
  year={2024}
}

@article{lewis2020retrieval,
  title={Retrieval-augmented generation for knowledge-intensive nlp tasks},
  author={Lewis, Patrick and Perez, Ethan and Piktus, Aleksandra and Petroni, Fabio and Karpukhin, Vladimir and Goyal, Naman and K{\"u}ttler, Heinrich and Lewis, Mike and Yih, Wen-tau and Rockt{\"a}schel, Tim and others},
  journal={NeurIPS},
  volume={33},
  pages={9459--9474},
  year={2020}
}

@inproceedings{chen2024benchmarking,
  title={Benchmarking large language models in retrieval-augmented generation},
  author={Chen, Jiawei and Lin, Hongyu and Han, Xianpei and Sun, Le},
  booktitle={AAAI},
  pages={17754--17762},
  year={2024}
}

@article{jiang2024mmsearch,
  title={Mmsearch: Benchmarking the potential of large models as multi-modal search engines},
  author={Jiang, Dongzhi and Zhang, Renrui and Guo, Ziyu and Wu, Yanmin and Lei, Jiayi and Qiu, Pengshuo and Lu, Pan and Chen, Zehui and Fu, Chaoyou and Song, Guanglu and others},
  journal={arXiv preprint arXiv:2409.12959},
  year={2024}
}

@article{meng2024simpo,
  title={Simpo: Simple preference optimization with a reference-free reward},
  author={Meng, Yu and Xia, Mengzhou and Chen, Danqi},
  journal={NeurIPS},
  volume={37},
  pages={124198--124235},
  year={2024}
}

@article{jiang2025deepretrievalhackingrealsearch,
      title={DeepRetrieval: Hacking Real Search Engines and Retrievers with Large Language Models via Reinforcement Learning}, 
      author={Pengcheng Jiang and Jiacheng Lin and Lang Cao and Runchu Tian and SeongKu Kang and Zifeng Wang and Jimeng Sun and Jiawei Han},
      year={2025},
      journal = {arXiv preprint arXiv: 2503.00223},
      url={https://arxiv.org/abs/2503.00223}
  }

@article{li2025search,
  title={Search-o1: Agentic search-enhanced large reasoning models},
  author={Li, Xiaoxi and Dong, Guanting and Jin, Jiajie and Zhang, Yuyao and Zhou, Yujia and Zhu, Yutao and Zhang, Peitian and Dou, Zhicheng},
  journal={arXiv preprint arXiv:2501.05366},
  year={2025}
}

@misc{ragen,
      title={RAGEN: Understanding Self-Evolution in LLM Agents via Multi-Turn Reinforcement Learning}, 
      author={Zihan Wang and Kangrui Wang and Qineng Wang and Pingyue Zhang and Linjie Li and Zhengyuan Yang and Xing Jin and Kefan Yu and Minh Nhat Nguyen and Licheng Liu and Eli Gottlieb and Yiping Lu and Kyunghyun Cho and Jiajun Wu and Li Fei-Fei and Lijuan Wang and Yejin Choi and Manling Li},
      year={2025},
      eprint={2504.20073},
      archivePrefix={arXiv},
      primaryClass={cs.LG},
      url={https://arxiv.org/abs/2504.20073}, 
}

@misc{chen2025r1v,
  author       = {Chen, Liang and Li, Lei and Zhao, Haozhe and Song, Yifan and Vinci},
  title        = {R1-V: Reinforcing Super Generalization Ability in Vision-Language Models with Less Than \$3},
  howpublished = {\url{https://github.com/Deep-Agent/R1-V}},
  note         = {Accessed: 2025-02-02},
  year         = {2025}
}

@article{meng2025mm,
  title={MM-Eureka: Exploring Visual Aha Moment with Rule-based Large-scale Reinforcement Learning},
  author={Meng, Fanqing and Du, Lingxiao and Liu, Zongkai and Zhou, Zhixiang and Lu, Quanfeng and Fu, Daocheng and Shi, Botian and Wang, Wenhai and He, Junjun and Zhang, Kaipeng and others},
  journal={arXiv preprint arXiv:2503.07365},
  year={2025}
}

@article{liu2025visual,
  title={Visual-RFT: Visual Reinforcement Fine-Tuning},
  author={Liu, Ziyu and Sun, Zeyi and Zang, Yuhang and Dong, Xiaoyi and Cao, Yuhang and Duan, Haodong and Lin, Dahua and Wang, Jiaqi},
  journal={arXiv preprint arXiv:2503.01785},
  year={2025}
}

@inproceedings{guu2020retrieval,
  title={Retrieval augmented language model pre-training},
  author={Guu, Kelvin and Lee, Kenton and Tung, Zora and Pasupat, Panupong and Chang, Mingwei},
  booktitle={ICML},
  pages={3929--3938},
  year={2020},
  organization={PMLR}
}

@article{yu2023augmentation,
  title={Augmentation-adapted retriever improves generalization of language models as generic plug-in},
  author={Yu, Zichun and Xiong, Chenyan and Yu, Shi and Liu, Zhiyuan},
  journal={arXiv preprint arXiv:2305.17331},
  year={2023}
}

@article{gao2023retrieval,
  title={Retrieval-augmented generation for large language models: A survey},
  author={Gao, Yunfan and Xiong, Yun and Gao, Xinyu and Jia, Kangxiang and Pan, Jinliu and Bi, Yuxi and Dai, Yixin and Sun, Jiawei and Wang, Haofen and Wang, Haofen},
  journal={arXiv preprint arXiv:2312.10997},
  volume={2},
  number={1},
  year={2023}
}

@article{bai2025qwen25,
  title={Qwen2. 5-vl technical report},
  author={Bai, Shuai and Chen, Keqin and Liu, Xuejing and Wang, Jialin and Ge, Wenbin and Song, Sibo and Dang, Kai and Wang, Peng and Wang, Shijie and Tang, Jun and others},
  journal={arXiv preprint arXiv:2502.13923},
  year={2025}
}

@article{hurst2024gpt,
  title={Gpt-4o system card},
  author={Hurst, Aaron and Lerer, Adam and Goucher, Adam P and Perelman, Adam and Ramesh, Aditya and Clark, Aidan and Ostrow, AJ and Welihinda, Akila and Hayes, Alan and Radford, Alec and others},
  journal={arXiv preprint arXiv:2410.21276},
  year={2024}
}

@article{wang2025vrag,
  title={VRAG-RL: Empower Vision-Perception-Based RAG for Visually Rich Information Understanding via Iterative Reasoning with Reinforcement Learning},
  author={Wang, Qiuchen and Ding, Ruixue and Zeng, Yu and Chen, Zehui and Chen, Lin and Wang, Shihang and Xie, Pengjun and Huang, Fei and Zhao, Feng},
  journal={arXiv preprint arXiv:2505.22019},
  year={2025}
}

@inproceedings{yao2022react,
  title={React: Synergizing reasoning and acting in language models},
  author={Yao, Shunyu and Zhao, Jeffrey and Yu, Dian and Du, Nan and Shafran, Izhak and Narasimhan, Karthik R and Cao, Yuan},
  booktitle={ICLR},
  year={2022}
}

@article{narayan2025deepmmsearch,
  title={Deepmmsearch-r1: Empowering multimodal llms in multimodal web search},
  author={Narayan, Kartik and Xu, Yang and Cao, Tian and Nerella, Kavya and Patel, Vishal M and Shiee, Navid and Grasch, Peter and Jia, Chao and Yang, Yinfei and Gan, Zhe},
  journal={arXiv preprint arXiv:2510.12801},
  year={2025}
}

@article{bai2025qwen3,
  title={Qwen3-vl technical report},
  author={Bai, Shuai and Cai, Yuxuan and Chen, Ruizhe and Chen, Keqin and Chen, Xionghui and Cheng, Zesen and Deng, Lianghao and Ding, Wei and Gao, Chang and Ge, Chunjiang and others},
  journal={arXiv preprint arXiv:2511.21631},
  year={2025}
}

@article{niu2025mineru2,
  title={Mineru2. 5: A decoupled vision-language model for efficient high-resolution document parsing},
  author={Niu, Junbo and Liu, Zheng and Gu, Zhuangcheng and Wang, Bin and Ouyang, Linke and Zhao, Zhiyuan and Chu, Tao and He, Tianyao and Wu, Fan and Zhang, Qintong and others},
  journal={arXiv preprint arXiv:2509.22186},
  year={2025}
}

@misc{shen2025rightwayassessingdocument,
      title={Are We on the Right Way for Assessing Document Retrieval-Augmented Generation?}, 
      author={Wenxuan Shen and Mingjia Wang and Yaochen Wang and Dongping Chen and Junjie Yang and Yao Wan and Weiwei Lin},
      year={2025},
      eprint={2508.03644},
      archivePrefix={arXiv},
      primaryClass={cs.CL},
      url={https://arxiv.org/abs/2508.03644}, 
}

@misc{chen2025visrbenchempiricalstudyvisual,
      title={VisR-Bench: An Empirical Study on Visual Retrieval-Augmented Generation for Multilingual Long Document Understanding}, 
      author={Jian Chen and Ming Li and Jihyung Kil and Chenguang Wang and Tong Yu and Ryan Rossi and Tianyi Zhou and Changyou Chen and Ruiyi Zhang},
      year={2025},
      eprint={2508.07493},
      archivePrefix={arXiv},
      primaryClass={cs.CV},
      url={https://arxiv.org/abs/2508.07493}, 
}

@misc{vanlandeghem2023documentunderstandingdatasetevaluation,
      title={Document Understanding Dataset and Evaluation (DUDE)}, 
      author={Jordy Van Landeghem and Rubén Tito and Łukasz Borchmann and Michał Pietruszka and Paweł Józiak and Rafał Powalski and Dawid Jurkiewicz and Mickaël Coustaty and Bertrand Ackaert and Ernest Valveny and Matthew Blaschko and Sien Moens and Tomasz Stanisławek},
      year={2023},
      eprint={2305.08455},
      archivePrefix={arXiv},
      primaryClass={cs.CV},
      url={https://arxiv.org/abs/2305.08455}, 
}

@misc{zou2024docbenchbenchmarkevaluatingllmbased,
      title={DOCBENCH: A Benchmark for Evaluating LLM-based Document Reading Systems}, 
      author={Anni Zou and Wenhao Yu and Hongming Zhang and Kaixin Ma and Deng Cai and Zhuosheng Zhang and Hai Zhao and Dong Yu},
      year={2024},
      eprint={2407.10701},
      archivePrefix={arXiv},
      primaryClass={cs.CL},
      url={https://arxiv.org/abs/2407.10701}, 
}
\bibliographystyle{colm2024_conference}
}

\newpage
\appendix

\section{Training Data Synthesis}

\subsection{Data Collection}
\label{app:data_collection}
To construct a diverse and comprehensive training corpus, we aggregate samples from several public benchmarks spanning a wide range of document types, languages, and reasoning requirements:

\begin{itemize}[leftmargin=*,noitemsep]
    \item \textbf{SlideVQA}~\citep{tanaka2023slidevqa} is a multi-image document VQA dataset containing over 2,600 slide decks, 52K images, and 14.5K QA pairs. It requires cross-slide information integration for single-hop, multi-hop, and numerical reasoning.
    
    \item \textbf{DoubleBench}~\citep{shen2025rightwayassessingdocument} is a large-scale multilingual multimodal document RAG dataset comprising 3,276 documents (72,880 pages) and 5,168 single-hop and multi-hop queries across 6 languages, covering PDFs, scanned documents, slides, and HTML pages.
    
    \item \textbf{VisRAG-Bench}~\citep{chen2025visrbenchempiricalstudyvisual} is a multilingual multimodal document retrieval dataset built from Common Crawl PDFs. It contains over 1,200 documents and 35,000 QA pairs across 16 languages, covering chart, text, and table-based questions.
    
    \item \textbf{DUDE}~\citep{vanlandeghem2023documentunderstandingdatasetevaluation} is a multi-domain, multi-page document understanding dataset spanning healthcare, legal, technical, and financial sectors. It includes both digitally-born and scanned documents with diverse layouts, time spans, and question types, simulating realistic document analysis scenarios.
    
    \item \textbf{DocBench}~\citep{zou2024docbenchbenchmarkevaluatingllmbased} is a benchmark for evaluating LLM-based document reading systems, containing 229 real-world documents and 1,102 QA pairs across academic, financial, government, legal, and news domains. QA pairs are generated via human annotation combined with GPT-4 synthesis, and categorized into text-based, multimodal, metadata, and unanswerable types.
\end{itemize}

\subsection{Automated Trajectory Synthesis.}
\label{sec:data_gen}
To acquire high-quality supervision that demonstrates effective coordination between retrieval, selection, and perception, we employ a strong proprietary LVLM (\textit{e.g.}, Qwen3-VL-235B~\citep{bai2025qwen3}) as a teacher agent, denoted as $\pi_{teacher}$, to synthesize reasoning trajectories. Formally, at each time step $t$, the teacher model observes the current history $H_{t-1}$ and generates the optimal thought-action pair:
\begin{equation}
(T_t, A_t) \sim \pi_{teacher}(\cdot|H_{t-1})
\end{equation}
While the generation of search and selection actions follows the standard interaction loop, the construction of \textit{Visual Perception} actions requires specific handling to ensure precision. 
When the teacher selects a relevant image $c_{sel}$ (via the \textit{Precise Selection} action), we introduce an intermediate layout analysis step. We utilize an expert document parsing tool, Mineru~\citep{niu2025mineru2}, to detect layout elements and generate a set of candidate bounding boxes $\mathcal{B}_{cand}$ representing potential regions of interest (ROIs). 
The teacher model $\pi_{teacher}$ then evaluates these candidates to determine if fine-grained perception is necessary. If a specific region contains critical information absent in the global view, the teacher selects the optimal bounding box $b^* \in \mathcal{B}_{cand}$ as the ground-truth crop action $A_{t+1}$. The cropped view is then processed as observation $O_{t+1}$, and the cycle continues.

\subsection{Data Filtering}
\label{app:data_filtering}
After performing trajectory synthesis as described in Section~\ref{sec:data_gen}, we apply a multi-stage filtering pipeline to ensure data quality:

\paragraph{Quality Filtering.}
We employ a strong teacher model (Qwen3-VL-235B) to verify the correctness of each synthesized trajectory by comparing the generated answer against the reference answer. Trajectories with incorrect final answers are discarded, ensuring that only factually accurate reasoning paths are retained for training.

\paragraph{Difficulty-Aware Filtering.}
To further reduce the proportion of trivially easy samples, we adopt a bootstrapped filtering strategy. We first fine-tune an intermediate SFT model using only the quality-filtered SlideVQA data, and then apply this model to generate answers for samples from the remaining datasets. Only those samples that (a) pass the quality filtering (i.e., the teacher-generated trajectory is correct) but (b) the intermediate SFT model answers incorrectly are retained. This procedure effectively removes samples that can already be solved by a partially trained model, yielding a more challenging and informative training set.

\paragraph{RL Data Curation.}
For the RL training set, we require samples that are challenging yet learnable. Specifically, we run the SFT model with high sampling temperature ($\tau$) for 5 independent rollouts per query. We then select samples where the model successfully retrieves relevant documents but fails to produce the correct answer in at least one rollout. This strategy identifies instances where the reasoning or perception steps—rather than retrieval—are the bottleneck. Additionally, we prioritize samples that were filtered out during the quality filtering stage, as they are likely to represent hard cases that maximally benefit from RL optimization.

\begin{table}[!t]
\small
\vspace{-1mm}
\centering

\vspace{-1mm}
\tabcolsep5pt

\setlength{\aboverulesep}{0pt}
\setlength{\belowrulesep}{0pt}
\renewcommand{\arraystretch}{1.2}

\scalebox{0.9}{
\begin{tabular}{c|cccccc}

    \toprule
    \rowcolor{CreamyBrown1} 
     &
    \multicolumn{6}{c}{\textsc{\textbf{Dataset}}} \\ 
    
    \rowcolor{CreamyBrown1} 
    \multirow{-2}{*}{\textsc{\textbf{Stage}}} & \textbf{SlideVQA} &\textbf{DoubleBench} &\textbf{VisR-Bench} &\textbf{DocBench} &\textbf{ DUDE} &\textbf{Total} \\

    \midrule
     SFT & 8271 & 1274 & 1518 & 657 & 901 & 12621\\
     RL & 2613 & 993 & 1630 & 301 & - & 5537\\
    \bottomrule
    \end{tabular}}
\vspace{-2mm}
\caption{\textbf{Data statistics for UniDoc-RL.}}
\vspace{-5mm}
\label{tab:data_stats}
\end{table}

\subsection{Data Statistics}
\label{app:data_statistics}
After the trajectory synthesis and multi-stage filtering pipeline, we obtain a final dataset comprising 12,621 samples for Supervised Fine-Tuning (SFT) and 5,537 samples for Reinforcement Learning (RL). The detailed composition across source datasets is presented in Table~\ref{tab:data_stats}.

\begin{figure*}[t!]
    \centering
    \includegraphics[width=\linewidth]{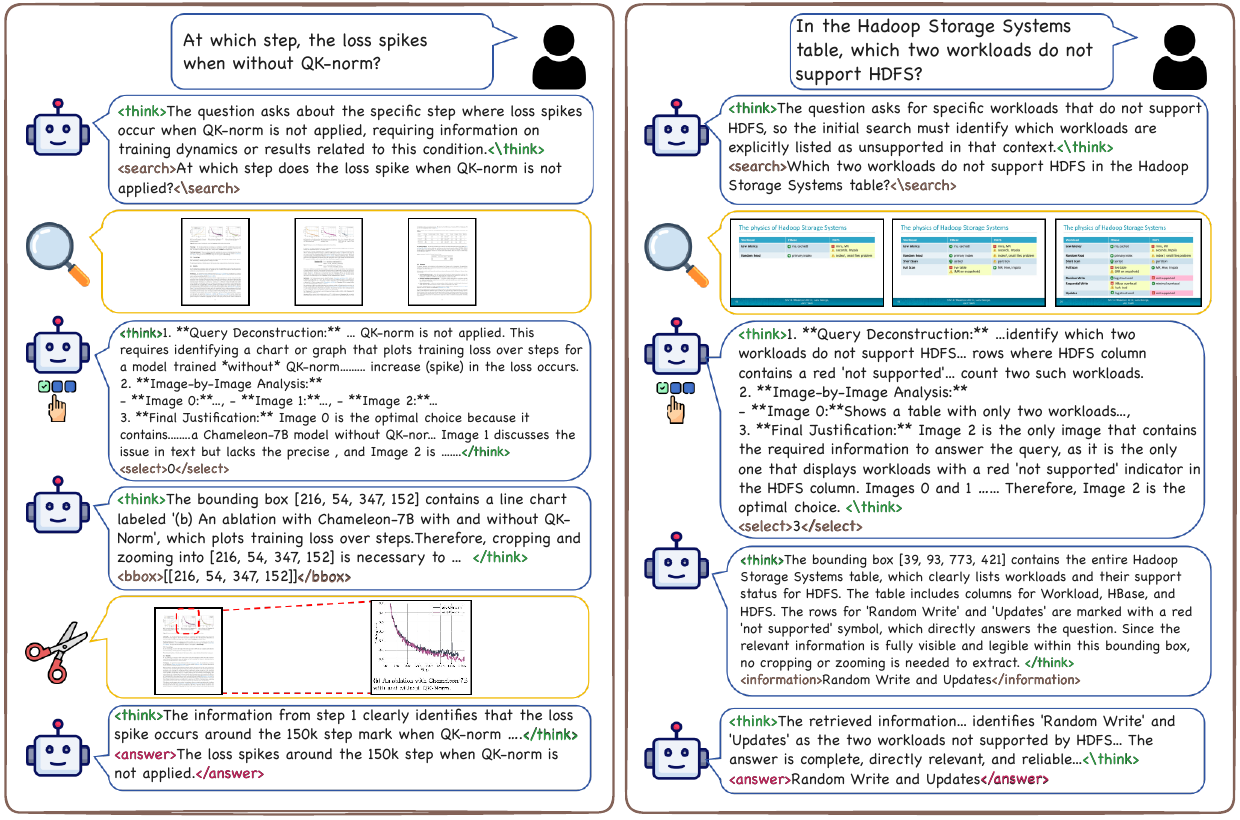}
    \vspace{-5mm}
    \caption{\textbf{Case Study for UniDoc-RL.}}
    \vspace{-7mm}
    \label{fig:case_study}
\end{figure*}

\section{Case Study}
Figure~\ref{fig:case_study} presents two representative examples that illustrate UniDoc-RL's adaptive reasoning process. In Case~1, the model retrieves candidate images, selects the one containing relevant training loss curves, and then actively crops a specific sub-region to zoom into a small chart which correctly identifies that the loss spikes around the 150k step. This demonstrates the necessity of the \textit{Visual Perception} action for extracting fine-grained details from complex figures. In Case~2, the model retrieves and selects a table that is already clearly legible, and intelligently skips the crop action, directly extracting the answer. These two cases jointly show that UniDoc-RL learns an adaptive strategy through RL training.

\section{Hyperparameters}
The detailed hyperparameters we use during training are shown in Table~\ref{tab:SFT_hyper} and Table~\ref{tab:RL_hyper}.     

\begin{table}[h]
\centering
\begin{minipage}[t]{0.48\textwidth} 
\centering
\begin{tabular}{@{}c|c@{}}
\toprule
\textbf{Name} & \textbf{Value} \\ \midrule
Finetuning type & Full \\
Freeze vision tower & True \\
Freeze multi-modal projector & True \\
Freeze language model & False \\
Cutoff len & 40000 \\
Epochs & 3 \\
Batch size & 1 \\
Gradient accumulation steps & 2 \\
Learning rate & 1.0e-5 \\
LR scheduler type & cosine \\
Warmup ratio & 0.1 \\ 
\bottomrule
\end{tabular}
\vspace{-2mm}
\caption{\textbf{Key hyperparameters for SFT.}}
\label{tab:SFT_hyper}
\end{minipage}
\hfill 
\begin{minipage}[t]{0.48\textwidth} 
\centering


\begin{tabular}{@{}c|c@{}}
\toprule
\tabcolsep5pt
\textbf{Name} & \textbf{Value} \\ \midrule
Number of agent groups & 5 \\
Warmup steps ratio & 0.285 \\
Mini batch size & 16 \\
Micro batch size per GPU & 1 \\
Learning rate (Actor) & 1.0e-6 \\
KL loss coefficient & 0.01 \\
Tensor model parallel size & 4 \\
Total epochs & 1 \\
Max prompt length & 40000 \\
Max response length & 1024 \\
$\lambda_{1:5}$ & 0.1,0.1,...,0.6 \\
\bottomrule
\end{tabular}
\vspace{-2mm}
\caption{\textbf{Key hyperparameters for RL.}}
\label{tab:RL_hyper}
\end{minipage}

\end{table}

\section{Model-Based Reward}
\begin{figure*}[t!]
    \centering
    \includegraphics[width=0.98\linewidth]{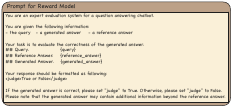}
    \vspace{-3mm}
    \caption{\textbf{Prompt for Reward Model.}}
    \vspace{-3mm}
    \label{fig:reward_prompt}
\end{figure*}

We employ a model-based reward to evaluate the quality and relevance of generated responses. Specifically, we utilize Qwen2.5-72B-Instruct~\citep{Qwen2.5-VL} as our reward model. The prompt used for the reward model is illustrated in Figure~\ref{fig:reward_prompt}. Given the input query, reference answer, and generated response, the reward model assesses the correctness of the generated response and outputs a binary value (0 or 1) to represent the accuracy of the answer.

\section{Prompts}

\begin{figure*}[t!]
    \centering
    \includegraphics[width=0.98\linewidth]{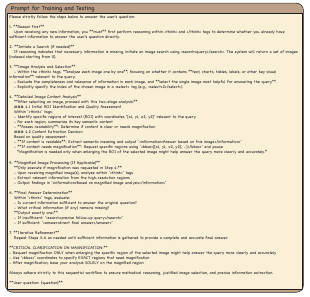}
    \vspace{-3mm}
    \caption{\textbf{Prompt for Training and Testing.}}
    \vspace{-5mm}
    \label{fig:prompt_train}
\end{figure*}

\begin{figure}[h]
    \centering
    \begin{minipage}[t][8cm][t]{0.48\textwidth}
        \centering
        \includegraphics[width=\textwidth]{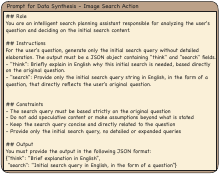}
        \vspace{-5mm}
        \caption{\textbf{Prompt for Data Synthesis - Image Search Action.}}
        \vspace{-7mm}
        \label{fig:prompt_search}
    \end{minipage}
    \hfill
    \begin{minipage}[t][8cm][t]{0.48\textwidth}
        \centering
        \includegraphics[width=\textwidth]{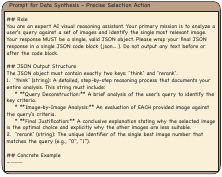}
        \vspace{-5mm}
        \caption{\textbf{Prompt for Data Synthesis - Precise Selection Action.}}
        \vspace{-7mm}
        \label{fig:prompt_selection}
    \end{minipage}
\end{figure}

\begin{figure*}[t!]
    \centering
    \includegraphics[width=0.98\linewidth]{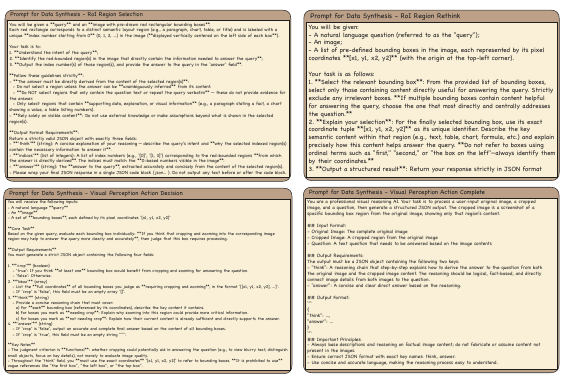}
    \vspace{-5mm}
    \caption{\textbf{Prompt for Data Synthesis - Visual Perception Action.}}
    \vspace{-3mm}
    \label{fig:prompt_vp}
\end{figure*}

We present the prompts for training and testing in Figure~\ref{fig:prompt_train} and the detailed prompts for data synthesis in Figures~\ref{fig:prompt_search},~\ref{fig:prompt_selection}, and~\ref{fig:prompt_vp}.

\end{document}